\pdfoutput=1

\documentclass[11pt]{article}

\usepackage[final]{acl}

\usepackage{times}
\usepackage{latexsym}

\usepackage[T1]{fontenc}

\usepackage[utf8]{inputenc}

\usepackage{microtype}

\usepackage{inconsolata}

\usepackage{amsmath,amsfonts,bm}

\def\eqref#1{equation~\ref{#1}}

\def\1{\bm{1}}

\DeclareMathAlphabet{\mathsfit}{\encodingdefault}{\sfdefault}{m}{sl}
\SetMathAlphabet{\mathsfit}{bold}{\encodingdefault}{\sfdefault}{bx}{n}

\DeclareMathOperator*{\argmax}{arg\,max}

\usepackage{graphicx}
\definecolor{topcolor}{rgb}{1,0.8,0.8}
\definecolor{secondcolor}{rgb}{1,0.87,0.7}
\definecolor{thirdcolor}{rgb}{1,1,0.8}
\usepackage{tikz}
\usepackage{pgfplots}
\usepackage[export]{adjustbox}
\usepackage{caption}
\usepackage{subcaption}
\usepackage{tabularx}
\definecolor{promptcolor}{RGB}{200, 235, 227}
\definecolor{promptcolor2}{RGB}{187,227,252}
\definecolor{prompttitlecolor}{RGB}{175, 172, 172}
\usepackage[framemethod=TikZ]{mdframed} 
\usepackage{wrapfig}
\usepackage{multirow}
\usepackage{graphicx}
\usepackage{booktabs}
\usepackage{colortbl} 
\mdfdefinestyle{prompt}{%
    linecolor =   black,
    linewidth =   0.2pt,
    backgroundcolor =  promptcolor2,
    roundcorner     =   1pt,
    font = \small\ttfamily,
    innerleftmargin=1.5pt,
    innerrightmargin=1.5pt
}

\newmdenv[%
    roundcorner=5pt, 
    linecolor =   black,
    linewidth =   1pt,
    font = \small\ttfamily,
    subtitlebackgroundcolor=prompttitlecolor, 
    frametitlebackgroundcolor=prompttitlecolor,
    backgroundcolor=promptcolor, 
    frametitle={Generated caption},
    subtitleaboveskip=0.5\baselineskip,
    subtitlebelowskip=0.5\baselineskip,
    ]{captionenv}

\DeclareMathAlphabet{\pazocal}{OMS}{zplm}{m}{n}

\DeclareMathAlphabet\mathbfcal{OMS}{cmsy}{b}{n}

\title{Local Prompt Optimization}

\author{\hspace{0.2cm}Yash Jain \qquad  Vishal Chowdhary \\ \\ \hspace{-1cm}Microsoft}

\begin{document}
\maketitle
\begin{abstract}
In recent years, the use of prompts to guide the output of Large Language Models have increased dramatically. However, even the best of experts struggle to choose the correct words to stitch up a prompt for the desired task. To solve this, LLM driven prompt optimization emerged as an important problem. Existing prompt optimization methods optimize a prompt globally, where in all the prompt tokens have to be optimized over a large vocabulary while solving a complex task.
The large optimization space (tokens) leads to insufficient guidance for a better prompt. In this work, we introduce Local Prompt Optimization (LPO) that integrates with any general automatic prompt engineering method. We identify the optimization tokens in a prompt and nudge the LLM to focus only on those tokens in its optimization step. We observe remarkable performance improvements on Math Reasoning (GSM8k and MultiArith) and BIG-bench Hard benchmarks across various automatic prompt engineering methods. Further, we show that LPO converges to the optimal prompt faster than global methods.
\end{abstract}

\section{Introduction}



Large Language Models (LLMs) are everywhere. LLMs are automating all the tasks that required specialized models a few years ago \citep{dubey2024llama3herdmodels, OpenAI2023GPT4TR}. The easiest and cheapest way to control an LLM's output is to do prompt engineering~\citep{pmlr-v202-zhou23g, Zhao2021CalibrateBU, yang-etal-2023-fantastic, lu-etal-2022-fantastically}. Unfortunately, writing a prompt is extremely tricky~\citep{pryzant-etal-2022-automatic}. Although the prompts are in English, the choice of words that effectively have the same meaning makes a huge difference in the prompt's performance on a task ~\citep{kojima2022large, wei2022cot, amatriain2024prompt}. Furthermore, an LLM is inherently biased towards its own vocabulary, making the task even more challenging. Thus, LLMs are used to modify prompts in a process called Prompt Optimization ~\citep{pmlr-v202-zhou23g}.

\begin{figure}[!t]
    \hspace{-6mm}\includegraphics[width=0.6\textwidth]{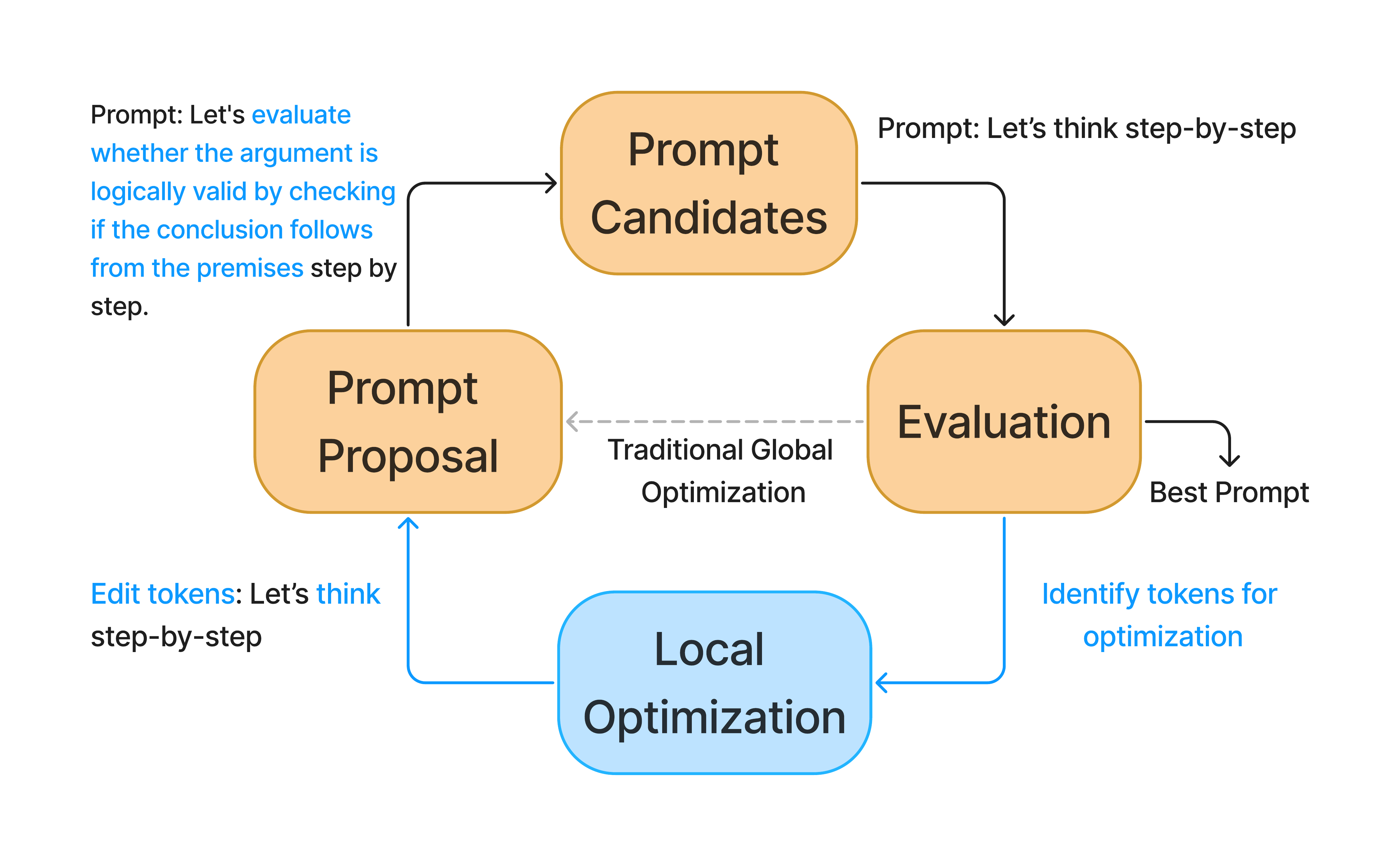}

    \vspace{-2mm}
    \caption{Local Prompt Optimization integrated in a general automatic prompt engineering framework.}
    \label{fig:main_fig}
    \vspace{-2mm}
\end{figure}

Prompt optimization techniques follow a two-step process as shown in Fig.~\ref{fig:main_fig}. First, the prompt is validated against a training set where the incorrect predictions are identified. Optionally, a feedback step is added where a natural language feedback, termed `textual gradients', is obtained by querying the LLM ~\citep{pe2, gpo}. Finally, the prompt is optimized using the textual gradients (incorrect examples or natural language feedback) to obtain an optimized prompt. The cycle is repeated for a fixed number of steps.

Traditional prompt optimization techniques ~\citep{apo, ape, pe2, gpo} optimize prompts globally \textit{,i.e.,} mutate all tokens in the prompt. However, optimizing all the tokens in a prompt while searching over the vocabulary to solve a complex problem, makes the prompt optimization very challenging. Further, for many production applications, it is desirable to optimize only subsections of the prompt while keeping the other parts static. Doing so requires us to limit the scope of the `prompt proposal' on subsection of the prompt, hence, the need of Local Prompt Optimization (LPO). Thus, we reduce the optimization space (tokens) for the LLM to simplify the problem and control the edit direction of a prompt. 

In this work, we evaluate the efficacy and pitfalls of doing local prompt optimization compared to global prompt optimization. We incorporate local optimization in three automatic prompt optimization algorithms and evaluate on GSM8k~\citep{gsm8k}, MultiArith~\citep{multiarith}, and BIG-bench hard ~\citep{bbh} benchmarks. We highlight that local optimization leads to faster convergence of optimal prompt while improving prompt performance. Finally, we test local optimization on a real-world application by evaluating it on a production prompt. 



\section{Background and Method}
\label{sec:method}
In this section, we will describe a general framework of automatic prompt engineering ~\citep{pmlr-v202-zhou23g} and highlight the gap in the framework. Building on this, we will introduce local prompt optimization.

\subsection{Automatic Prompt Engineering}
Given a dataset $D = {(x,y)}$, a prompt engineering system aims to find a prompt $p^*$ that maximizes the score on an evaluator function $f$. Specifically, 
\begin{equation}
     p^* = \argmax_{p} \sum_{(x,y)\in D} f(\mathcal{M}_{task}(x; p), y)\label{eq:pe}
\end{equation}
where $\mathcal{M}_{task}(x; p)$ is the output generated by the task model $\mathcal{M}_{task}$ when conditioning on the prompt $p$.

A general automatic prompt engineering system has three parts: Prompt Initialization, Prompt Proposal, and Search Procedure.
\paragraph{(1) Prompt Initialization:} An initial prompt is provided to an automatic prompt system that needs to be optimized. Prompt Initialization can be done by a manual human-written instruction or it can be few shot examples from the dataset $D$ ~\citep{Zhao2021CalibrateBU}. 
\paragraph{(2) Prompt Proposal:} In this step new prompt generation takes place. At any timestep $t$, a new set of prompts $p^{(t+1)}$ are generated from a set of candidate prompts $p^t$. A proposal LLM $\mathcal{M}_{proposal}$ is used to propose new prompts, grounded on `textual gradients' $g^t$ obtained on the current prompt $p^t$. These `textual gradients' consists of a meta prompt along with additional information which vary between automatic prompt engineering techniques. These include  incorrect examples ~\citep{ape}, or a natural language LLM feedback of the incorrect examples ~\citep{apo} to a combination of both along with previous prompts $p^{(t-1)}$ and their scores ~\citep{pe2}. 
\begin{equation}
    p^{(t+1)} = \mathcal{M}_{proposal}(p^t, g^t).
\end{equation}
However, the edits in prompt $p^{(t)}$ can take place anywhere inside the prompt including complete re-writing the prompt at every timestep causing slow update towards the optimal prompt. Further, it does not provide any control required in a typical production prompt engineering where a professional would want prompt edits to take place within a specific scope of the prompt. Thus, the global optimization leads to slow prompt convergence and provides no control over direction of prompt optimization.

\paragraph{(3) Search:} Finally, among the candidate prompts across all timesteps $p^0 \cup p^1 \cup ... \cup p^t$, a subset of the best performing prompts are retained and the process is repeated.

\subsection{Local Prompt Optimization}
The basic function of `textual gradients` $g^t$ is to inform how the optimization process (gradient values) should adjust according to model's performance ~\citep{gpo}. However, it does not specify where the optimization should take place or analogously in deep learning on which parameters should the gradient descent should take place. We incorporate this intuition of parameter selection to reduce the optimization space through local prompt optimization.

Following the intuition of Chain-of-Thought logic~\citep{wei2022cot}, we first identify the potential tokens in the prompts which are responsible for incorrect predictions by adding an instruction in the meta-prompt before the Prompt Proposal step as depicted in Fig.~\ref{fig:main_fig}. We use \texttt{<edit>} tags to highlight the edit tokens, the meta-instruction is shown in Fig.~\ref{fig:lpo_prompt}. The goal is to identify tokens within the prompt that the proposal LLM $\mathcal{M}_{proposal}$ should optimize.

Once the prompt edit tokens are identified, we proceed with the Prompt Proposal step. The instruction `\texttt{Reply with the new instruction without the <edit>, </edit> tags.}' is provided to $\mathcal{M}_{proposal}$ to output the updated prompt $p^{(t+1)}$. Tab~\ref{tab:math_prompts} shows the complete prompt evolution with local and global optimization.




\begin{figure}[!ht]

\begin{mdframed}[style=prompt]

\small
First, identify the scope of tokens within the prompt where edits should take place. \\
Prompt edits include adding, deleting or modifying tokens. \\
Mark the scope of the prompt that needs editing by putting <edit>, </edit> tags. \\
You can have multiple <edit> tags and each <edit> tag should not entail more than 5 words. \\
Do not cover the whole sentence with multiple <edit> tags.\\
Reply with the prompt with <edit>, </edit> tags. \\
Do not include any other text.
\end{mdframed}
\caption{Illustration of the Prompt for identifying potential optimization tokens.}
\label{gpt-4-text}
\vspace{-3mm}
\label{fig:lpo_prompt}
\end{figure}

\section{Experiments}
The goal of this section is to highlight the efficacy of local optimization over existing global optimization across different automatic prompt engineering methods. 

\subsection{Datasets}
Following PE2~\citep{pe2} closely, we perform evaluation on three set of tasks varying in their objectives and domain. We use the same train-dev-test split as provided by ~\citep{pe2}.
\paragraph{(1) BIG-bench Hard} or BBH ~\citep{bbh} is a set of 23 tasks (27 subtasks) which can be categorized as algorithmic, natural language understanding, world knowledge, and multlingual reasoning tasks.
\paragraph{(2) Math Reasoning} consists of two datasets MultiArith ~\citep{multiarith} and GSM8K ~\citep{gsm8k}. Both contains grade school math problems requiring 2 to 8 steps of algebraic reasoning to reach the final answer. 
\paragraph{(3) Production Prompt} is an internal classification prompt, developed to orchestrate the correct tool for further LLM calls. The prompt would take in a user query and would identify the `intent' of the query. It would then output a function call with appropriate arguments. It has been developed by in-domain experts and is 8k tokens long. The prompt contains sections of skill definitions, specific classification instruction, safety instructions and so on, making it an ideal candidate for evaluation. 

\subsection{Prompt Optimization methods}
For fair comparison, we select three representative prompt optimization techniques and modify their global optimization step with our local optimization step as explained in Sec.~\ref{sec:method} and Fig.~\ref{fig:main_fig}. (1) \textbf{APE} ~\citep{ape} leverages LLMs to come up with variants of the input prompt, given few examples and then select the best performing prompt. An improved variant of APE called \textbf{Iterative APE}, repeats this process a few times to get a better optimized prompt. We use Iterative APE for comparison in the paper. (2) \textbf{APO} ~\citep{apo} is builds over Iterative APE and adds an incorrect prediction feedback in its prompt optimization process. This feedback is often termed as `textual gradients' and is used to make edits in correct direction on the candidate prompt. \textbf{APO} is named as \textbf{ProTeGi} in their recent draft. (3) \textbf{PE2} ~\citep{pe2} further innovates in the `textual gradients' and make them rich by adding old prompt and their feedback history to guide the edit process. They also limit the number of edits in the prompt.

\subsection{Implementation Details}
Across all experiments, we consistently use \texttt{gpt-3.5-turbo} as the task solving model and \texttt{gpt-4o} as the prompt optimizer. The remaining design details follow those of PE2~\citep{pe2}. We limit the search budget to 3 optimization steps, using a beam size of 4 and generating 4 prompts at each step. Further, we initialize the prompts for BBH and Math Reasoning datasets with the standard prompt ``Let's think step by step'' ~\citep{kojima2022large, wei2022cot}. We keep the hyperparameters for all the prompt optimization methods same across global and local optimization.

\begin{table}[t]
\centerline{
\scalebox{0.72}{
\begin{tabular}{l|p{0.5\textwidth}}
\toprule
Initial Prompt & Let's think step by step.    \\\midrule
\rowcolor{gray!20}\multicolumn{2}{l}{Global Optimization}    \\\midrule
\multirow{4}{*}{Optimum}& Ensure all given initial values and specific contexts (e.g., rounding rules, phrase interpretation) are considered, and explain the arithmetic operations logically and clearly, step-by-step.  \\\midrule
\rowcolor{gray!20}\multicolumn{2}{l}{Local Optimization}  \\\midrule
\multirow{2}{*}{Identifying edit}  & Let's \texttt{<edit>} think \texttt{</edit>} \texttt{<edit>} step by step \texttt{</edit>}. \\\midrule
\multirow{3}{*}{Optimum} & Let's carefully read and clearly understand the problem. Next, let's think through each step and verify each calculation carefully. \\
\bottomrule      
\end{tabular}
}}

\caption{MultiArith
prompts found by comparing traditional global optimization approach against our proposed local optimization.}\label{tab:math_prompts}
\end{table}

\begin{figure*}[ht]
    \centering
    \begin{subfigure}[b]{0.39\textwidth}
        \centering
        \includegraphics[width=\linewidth, height=4cm]{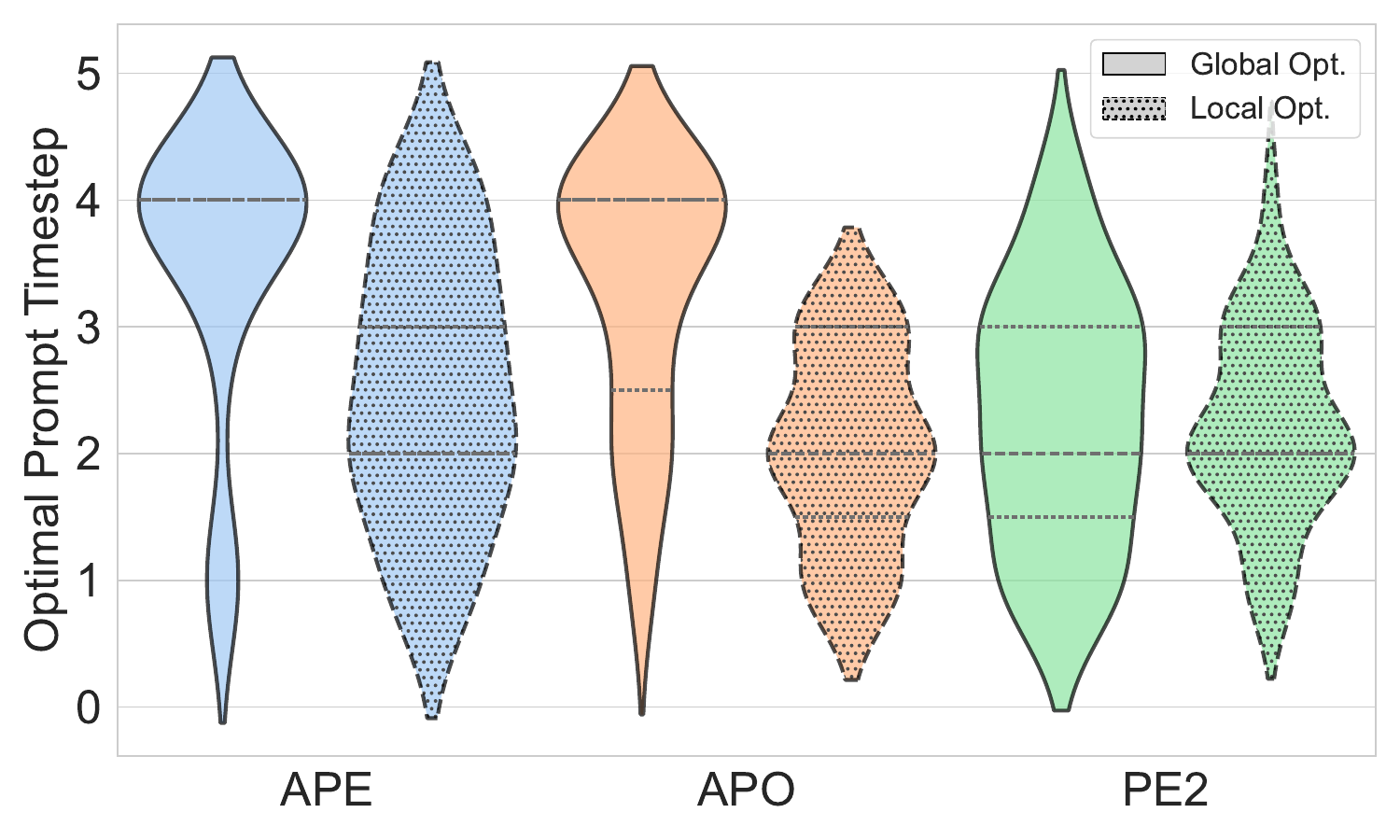}
        \caption{Optimal Prompt Timestep in the 27 subtasks of BBH benchmark. Local Opt. achieves faster convergence.}
        \label{fig:figure1}
    \end{subfigure}
    \hfill
    \begin{subfigure}[b]{0.39\textwidth}
        \centering
        \includegraphics[width=\linewidth, height=4cm]{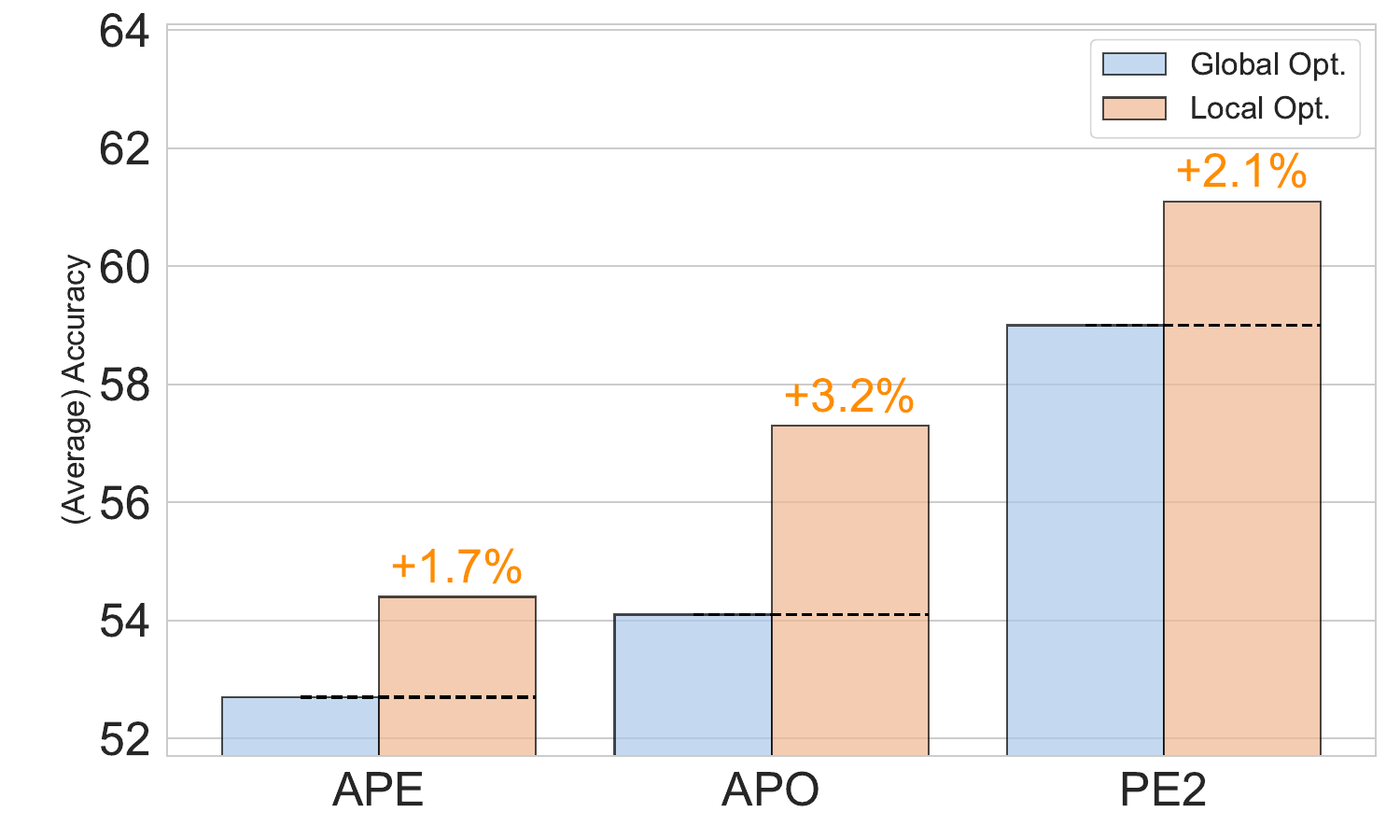}
        \caption{Average Accuracy on BBH. Local Opt. consistently outperforms global opt. across various methods.}
        \label{fig:figure2}
    \end{subfigure}
    \hfill
    \begin{subfigure}[b]{0.19\textwidth}
        \centering
        \includegraphics[width=\linewidth, height=4cm]{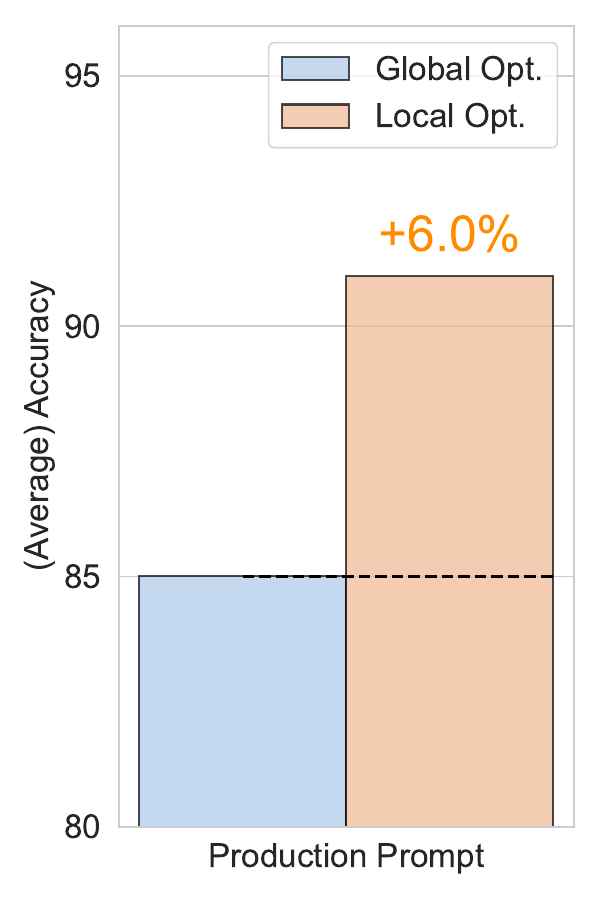}
        \caption{Production Prompt performance after employing local opt.}
        \label{fig:figure3}
    \end{subfigure}
    \caption{Experiments on BBH and Production Prompt, showcasing LPO benefits in both performance and efficiency.}
    \label{fig:threefigures}
\end{figure*}

\section{Results and Analysis}

\begin{table}[]
\centering
\resizebox{\columnwidth}{!}{%
\begin{tabular}{lcccc}
\toprule
Method & LPO & GSM8k ($\uparrow$) & MultiArith ($\uparrow$) & \# steps ($\downarrow$)     \\ \midrule
\multirow{2}{*}{APE} & - & 77.7 & 93.2   & \textbf{2.5} \\
                     &  \checkmark & \textbf{78.0} &\textbf{ 96.2}  & 4   \\ \midrule
\multirow{2}{*}{APO} & - & 77.7 & 96.0   & 4 \\
                     &  \checkmark &\textbf{ 79.7} & \textbf{97.5} & \textbf{2} \\ \midrule
\multirow{2}{*}{PE2} & - & 78.7 & 97.0   &  2.5 \\
                     &  \checkmark & \textbf{80.6} & \textbf{97.5 }& \textbf{2} \\
\bottomrule
\end{tabular}%
}
\caption{Results of Local Prompt Optimization (LPO) on Math Reasoning benchmark.}
\label{tab:my-table}
\end{table}


\paragraph{Local Prompt Optimization improves existing automatic prompting techniques.}
We evaluate APE, APO and PE2 algorithms with and without Local Optimization on GSM8K and MultiArith datasets as depicted in Tab.~\ref{tab:my-table}. We observe that Local Prompt Optimization is able to improve prompts for Math Reasoning tasks by an average of $1.5\%$ while decreasing the number of optimization steps required. Additionally, we demonstrate the wide applicability of Local optimization on BIG-bench Hard benchmark (27 subtasks). In Fig.~\ref{fig:figure2}, we show that local optimization supports various automatic prompting techniques over a large variety of tasks. We outperform traditional global optimization approach by an average of $2.3\%$ across methods. We hypothesize that since Local Optimization reduces the optimization tokens for the proposal LLM $\mathcal{M}_{proposal}$ and introduces a Chain-of-Thought approach in the optimization step, $\mathcal{M}_{proposal}$ is able to more efficiently solve the task and provide better prompt outputs. 

\paragraph{Local Prompt Optimization results in faster convergence.}
We estimate the timestep where the optimal prompt is produced over the 27 subtasks in BIG-bench Hard benchmark. The number of iterations were kept to 3 and we assign a timestep of 4 when the initialization prompt is found to be the best performing prompt.  Fig.~\ref{fig:figure1} depicts the violin curves of optimal prompt timestep. Notably, we observe majority of tasks reaching earlier convergence than global optimization approaches, saving a lot of LLM compute and time. Global optimization often leads to rewriting the complete prompt from scratch for the LLM, making the task more challenging and complex. On the other hand, we believe reducing the optimization space through local optimization keeps the gradient updates aligned towards the minima. 



\paragraph{Local Prompt Optimization can allow control over prompt editing.} 
Perhaps, the biggest benefit of LPO is to control the scope of editing. In the production prompt written by domain expert, the prompt has specific sections where the different tools are defined followed by instructions on individual tools and their use. Using LPO, we can specify which tool's instruction needs to be updated without affecting the other tools. Further, it ensures that there is no regression in performance of the prompt in other classes due to the optimization process. In our evaluation, we gained a significant jump of $6\%$ on the production prompt as shown in Fig.~\ref{fig:figure3}.

\section{Conclusion}
In this work, we identify the gap in the optimization step of the existing automatic prompt engineering techniques. Traditionally, prompts are mutated globally in each step. However, this global optimization increases the task complexity as the optimizer (LLM) has to work on a larger number of parameters (tokens) to find the optimal update. Furthermore, many production prompts require optimizing only a section of the prompt and not rewriting the complete prompt from scratch. As a fix, we introduce Local Prompt Optimization (LPO) where we identify the optimization tokens and nudge the optimizer to focus only on those tokens. We observe consistent performance improvements over Math Reasoning and BIG-bench Hard benchmark. Notably, we observe that local optimization searches the optimal prompt significantly quicker than the traditional approach. Further, LPO can be integrated well with long prompts, which are more common in practical settings, further showcasing the ubiquity of our method. Looking ahead, we are optimistic about prompt optimization techniques built from the perspective of local optimization to benefit from the gains in performance and efficiency.

\section{Limitations}
We believe our study has three limitations which we believe can be overcome in future works. (1) Multilinguality: We primarily focused on English language as the base in this work, from prompts to datasets to LLMs. However, we believe the ideas introduced in the paper are extendable to other languages as well and implore the community to build over our work. (2) Local Optimization sometimes leads to overfitting the prompt with dev score reaching close to $99\%$. We believe that a better search strategy can solve this problem and hope to see future works addressing it. (3) Closed-source models: We have used \texttt{GPT-4o} as the optimizer to benchmark large datasets in this work. This poses a challenge to the reproducibility of this work. However, we believe that showcasing local optimization capabilities on proprietary models is a good signal for both academic and industry to incorporate the ideas in their prompt engineering methods. 

\bibliography{main, acl_anthology}




\end{document}